\documentclass[10pt,twocolumn,letterpaper]{article}
\usepackage{cvpr}
\usepackage{times}
\usepackage{epsfig}
\usepackage{graphicx}
\usepackage{subcaption}
\usepackage{wrapfig}
\usepackage{amsmath}
\usepackage{amssymb}
\usepackage[inline]{enumitem} 

\usepackage[pagebackref=true,breaklinks=true,letterpaper=true,colorlinks,bookmarks=false]{hyperref}

\cvprfinalcopy 

\def\cvprPaperID{6024} 

\ifcvprfinal\fi
\begin{document}

\title{Re-Identification Supervised Texture Generation}

\author{Jian Wang$^{1}$\thanks{This work was done during Jian Wang, Yunshan Zhong and Yachun Li were interns at Megvii Technology.}, Yunshan Zhong$^{2*}$, Yachun Li$^{3*}$, Chi Zhang$^{4}$, and Yichen Wei$^{4}$\\
	\\
	$^1$State Key Lab. of Computer Science, ISCAS \& University of Chinese Academy of Sciences\\
	$^2$Peking University \quad $^3$Zhejiang University \quad	$^4$Megvii Technology\\
	{\tt\small wangj@ios.ac.cn, Zhongyunshan@pku.edu.cn, liyachun@zju.edu.cn}\\
	{\tt\small \{zhangchi,weiyichen\}@megvii.com}
}


\maketitle

\thispagestyle{empty}

\begin{abstract}
The estimation of 3D human body pose and shape from a single image has been extensively studied in recent years. However, the texture generation problem has not been fully discussed. In this paper, we propose an end-to-end learning strategy to generate textures of human bodies under the supervision of person re-identification. We render the synthetic images with textures extracted from the inputs and maximize the similarity between the rendered and input images by using the re-identification network as the perceptual metrics. Experiment results on pedestrian images show that our model can generate the texture from a single image and demonstrate that our textures are of higher quality than those generated by other available methods. Furthermore, we extend the application scope to other categories and explore the possible utilization of our generated textures.
\end{abstract}

\section{Introduction}\label{introduction}
\begin{figure}[!t]
	\centering
	\includegraphics[width=0.97\linewidth]{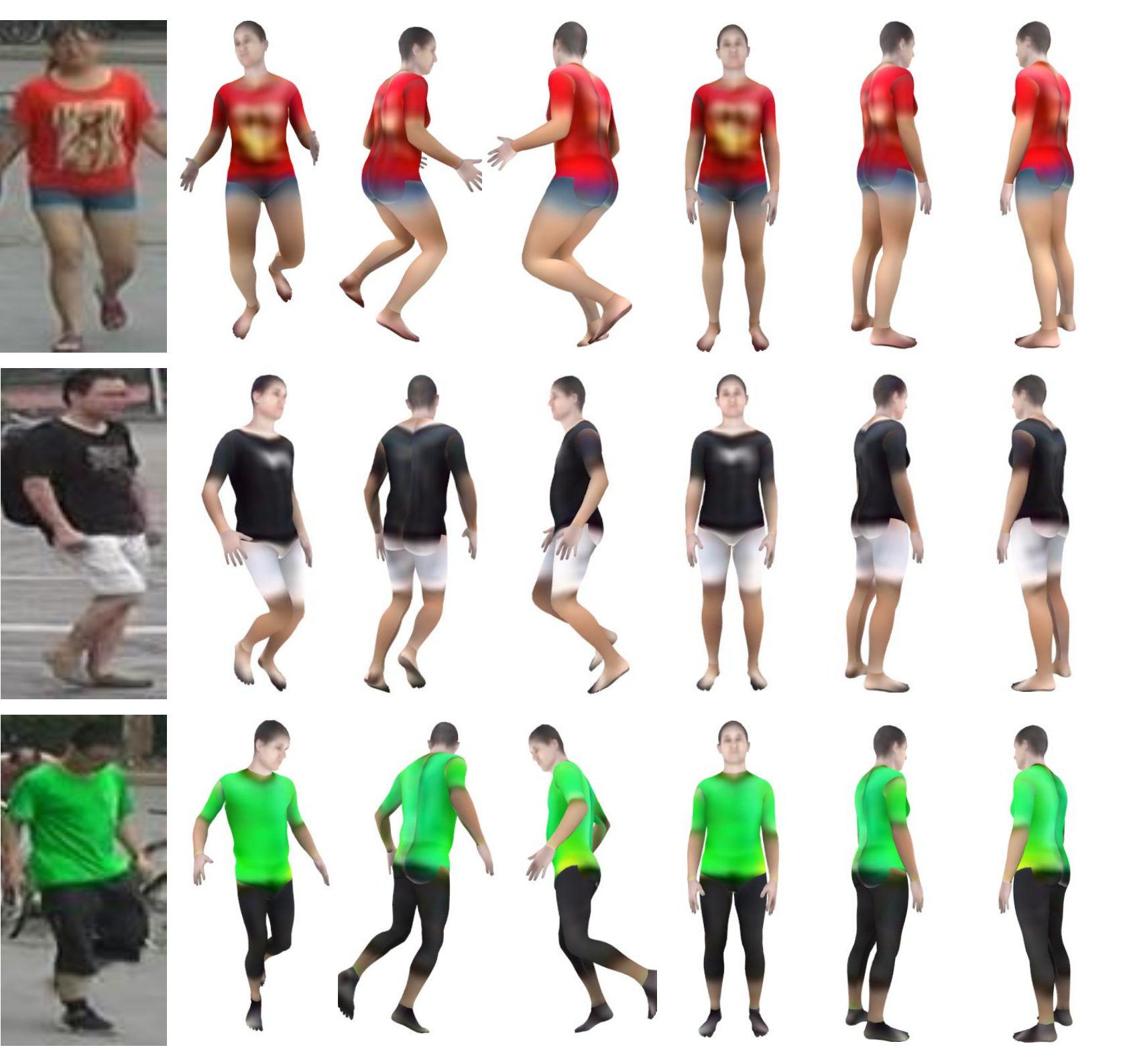}
	\caption{Texture generation results on Market-1501. This figure shows the original images (1st column), rendered 3D models in different view points (2nd-4th columns) and rendered 3D models in standing pose (5th - 7th columns). For better visual results, these 3D models are rendered with Blender \cite{blender}.}
	\label{demo}
\end{figure}

The automatic generation of realistic 3D human models is crucial for many applications, including virtual reality, animation, video editing and clothes try-on. Among the 3D human reconstruction approaches, generating the 3D human model from a single image receives more and more attention. A lot of methods have been proposed in both traditional \cite{unitethepeople,estimatinghumanshape,smplify} and deep learning manner \cite{end2end,neuralbodyfitting,bodynet}. Even though these methods succeed in estimating the pose and shape of the human body accurately, the generation of texture is omitted, which is the missing part in the reconstructing the realistic 3D human body.

Even though generating human body textures from a single image is of vital importance, there are only two methods which aim at solving it. \cite{neverova2018dense} first extract the partially observed textures from different images with DensePose \cite{guler2018densepose} and obtain the full textures by combining the partial ones. Then \cite{neverova2018dense} uses the full textures as ground truth and trains a generative network to directly infer the corresponding texture from a single image. This process is computationally expensive and requires high-quality image based dense human pose detection method for extracting the partially observed textures, which would be challenging for lots of in-the-wild images. \cite{DBLP:conf/eccv/KanazawaTEM18} renders synthetic images with generated textures and minimize the distance between the rendered image and the input image. However, \cite{DBLP:conf/eccv/KanazawaTEM18} uses an ImageNet-pretrained perceptual loss as the distance metric, which restricts the quality of their textures.

The shortcomings of existing works indicate that generating textures from a single image is challenging, which is caused by two reasons. Firstly, the occlusion caused by the human body makes it impossible to get the texture information from the occluded parts. Secondly, the diversity of human poses and the background complicates the texture extraction process. For example, the inaccuracy of available 3D pose estimation methods such as \cite{end2end} makes directly mapping the input images to 3D models difficult. 

To overcome these obstacles, we introduce the re-identification to supervise our texture generation model. Re-identification, the person identifying and retrieving method, is explicitly trained to minimize the distance between images from different viewpoints with the same identity. As the person identity is mainly characterized by textures, the re-identification network can serve as the distance metric for the textures partially observed from different viewpoints. This solves the first problem. Moreover, the re-identification network can extract the body features while eliminating the influence of pose and background variations \cite{zheng2016person}, which solves the second problem. From the reasons above, it can perform well as the supervision to guide the training process of texture generation networks.

Based on the supervision of re-identification, we propose a novel method to generate body textures from a single image. Example results are shown in Fig.~\ref{demo}. In order to train our model in an end-to-end way, we render images with the SMPL body model and use the distance between features extracted by re-identification network as the training loss (denoted as \emph{re-identification loss}). Our method shows the strong capability to efficiently generate body textures. 

In order to demonstrate the importance of re-identification network, we compare the re-identification loss with other loss functions which are commonly used in image generation tasks. Our experiments indicate that the performance of the model surpasses others in generating body textures. 

Aside from generating human body textures, we expand our method to other categories. Our method can generate better bird textures comparing with the approach presented in \cite{DBLP:conf/eccv/KanazawaTEM18}. In addition, the diversity of generated textures is higher than available 3D-scanned textures. The experiment on the action recognition task has demonstrated that it is beneficial to pretrain the network on dataset synthesized with highly diversified textures.

In summary, our contribution can be distributed into three aspects. Firstly, we introduce a new method to generate textures from a single image by incorporating the re-identification loss. Secondly, we provide an in-depth analysis to prove the effectiveness of re-identification loss in the texture generation task. Finally, we extend our method to broader object categories and explore the potential ability of our method as a data augmentation strategy.

\section{Related Work}

\paragraph{Texture generation.}
The texture generation is an essential task for reconstructing realistic 3D models because the texture represents crucial information for describing and identifying object instances. Most of the recent works focus on combining texture fragments from different views. \cite{highqualitytexture,maskedphotoblending,colouredsigneddistance} blend multiple images into textures with various weighted average strategies. However, these methods are sensitive to noises introduced by camera poses or 3D geometry and end up with blurring and ghosting. Some other methods \cite{lettherebecolor,perez2003poisson,allene2008seamless,gal2010seamless,lempitsky2007seamless} project images to appropriate vertices and faces. These approaches alleviated the blurring and ghosting problems while they are vulnerable to texture bleeding. Warping-based methods \cite{colormap,eisemann2008floating,aganj2009multi,bi2017patch} incorporate warping refinement technologies to correct for local misalignments. Specifically, \cite{videobased} apply back-project technology for 3D human body texture generation while \cite{detailedhuman} applies a semantic prior and graph optimization strategy to obtain finer details. These methods can build high-quality 3D textures, while images from different views or RGB-D sensors are required. 

Aside from the multi-view based texture generation, another challenging problem is generating textures from a single image. \cite{neverova2018dense} proposes a new pose transfer method which incorporates the human body texture generation module and \cite{DBLP:conf/eccv/KanazawaTEM18} proposed a method for recovering the textures of a specific category. Their approach is either computationally expensive or suffers from the low quality of generated textures, which is indicated in Sec.~\ref{introduction}.

\paragraph{Model-based 3D human pose and shape estimation.}
The model-based method estimates the human body pose and shape from a single image by fitting parameters of a specific body model. Earlier works like \cite{DBLP:conf/iccv/GuanWBB09,DBLP:conf/cvpr/BalanSBDH07} optimize the pose and shape parameters of SCAPE \cite{DBLP:journals/tog/AnguelovSKTRD05} under the supervision of human body key-points and silhouettes. However, SCAPE is not consistent with existing animation software, which limits its application scope.

Most of the recent works build their approaches on a simple yet powerful body model: SMPL (Skinned Multi-Person Linear Model) \cite{DBLP:journals/tog/LoperM0PB15}. SMPL renders the body mesh by calculating a linear function of pose and shape parameters, which enables the optimization of SMPL model by learning from massive data. \cite{smplify} designed a loss function to penalize the difference between projected 3D body joints and detected 2D joints. This loss function also prevents the inter-penetration between limbs and trunk. \cite{unitethepeople} infers the 3D shape and pose directly from 91 landmark predictions in UP-3D dataset, which accelerates the SMPL by one and two orders of magnitudes. \cite{videobased} and \cite{DBLP:conf/cvpr/BalanSBDH07} proposed a method to obtain a visual hull by transforming the silhouette cones corresponding to dynamic human silhouettes, which enables the accurate estimation of body shapes and textures. 

More and more recent work applies deep learning methods to fit SMPL parameters. \cite{DBLP:conf/bmvc/TanBC17,DBLP:conf/cvpr/DibraJOZG17,pavlakos2018learning} predict SMPL parameters directly by a deep neuron network and get supervision from differentiable rendering of silhouettes.  \cite{DBLP:conf/nips/TungTYF17} proposed a self-supervised method using 2D human keypoints, 3D mesh motions, and human-background segmentation. \cite{end2end} regresses the SMPL parameters iteratively and incorporates the prior knowledge to guarantee the reality of human shape and pose. \cite{neuralbodyfitting} process the image to 12 semantic segmentation parts and predict the SMPL parameters from them. \cite{bodynet} optimizes the volumetric loss to gain higher accuracy in body shape than previous methods.

\paragraph{Person re-identification.}
Person re-identification aims at spotting a person of interest in different cameras \cite{zheng2016person}. From the independence of person re-identification task in 2006 \cite{DBLP:conf/cvpr/GheissariSH06}, it has become increasingly popular due to its wide application. After the incorporation of CNN-based method in \cite{DBLP:conf/icpr/YiLLL14} and \cite{DBLP:conf/cvpr/LiZXW14}, the performance of person re-identification network is promoted drastically. For example, since the release of Market-1501 dataset \cite{DBLP:conf/iccv/ZhengSTWWT15} in 2015, the top-1 accuracy of state-of-the-art method has increased from 44.42\% in \cite{DBLP:conf/iccv/ZhengSTWWT15} to 96.6\% in \cite{wang2018learning}.

The core idea of deep learning in person re-identification is to extract features of the person from the image \cite{zheng2016person}. Moreover, the features of different body parts provide more fine-grained information than global features, thus combining local representations from parts of images has become one of the most prevalent strategies in recent works. For example, \cite{DBLP:conf/cvpr/ChengGZWZ16,DBLP:conf/ijcai/LiZG17,pcb,wang2018learning} split the image horizontally and learn local features in each part. \cite{DBLP:conf/cvpr/LiC0H17,DBLP:conf/iccv/ZhaoLZW17,li2018harmonious} apply region proposal methods to extract different human parts. Attention mechanism of \cite{DBLP:conf/iccv/LiuZTSSYYW17,li2018harmonious} shows priority in learning soft pixel-wise parts of the body. Based on recent advances, our approach utilizes the part-based person re-identification method to represent spatial features, which plays a key role in restoring detailed textures.

\section{Method}

\begin{figure}[h]
	\centering
	\includegraphics[width=0.97\linewidth]{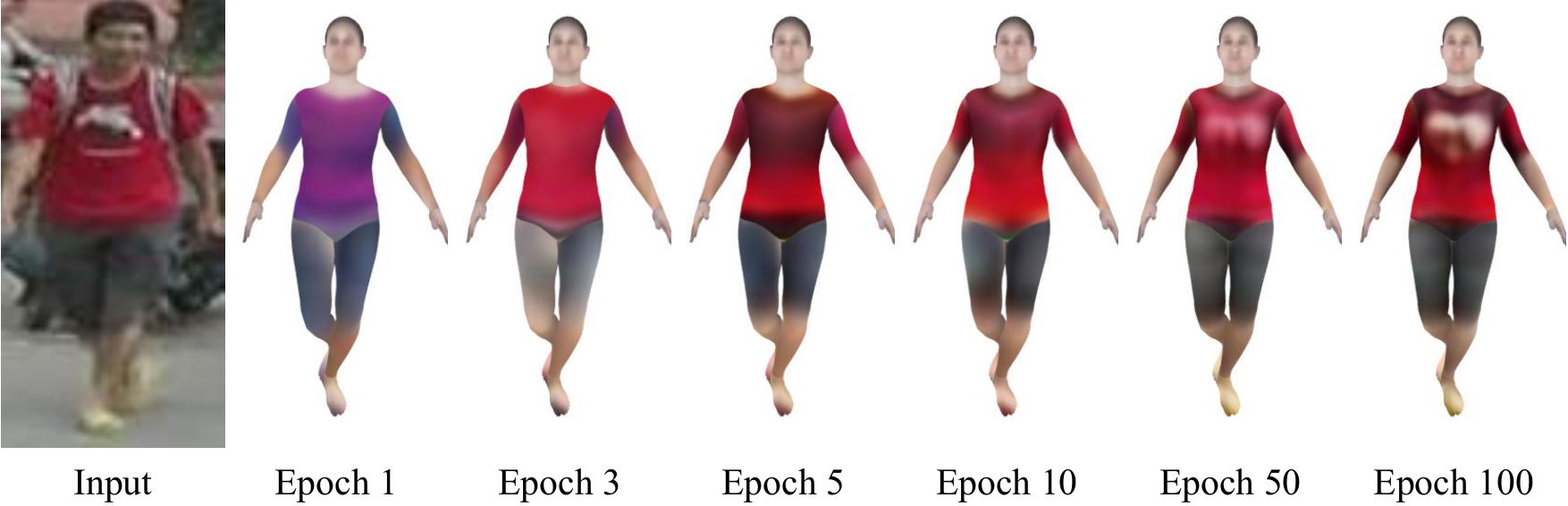}
	\caption{\textbf{Visualization of training process.} We render the image with textures generated in different training epochs.}
	\label{training}
\end{figure}

\begin{figure*}
	\centering
	\includegraphics[width=0.97\linewidth]{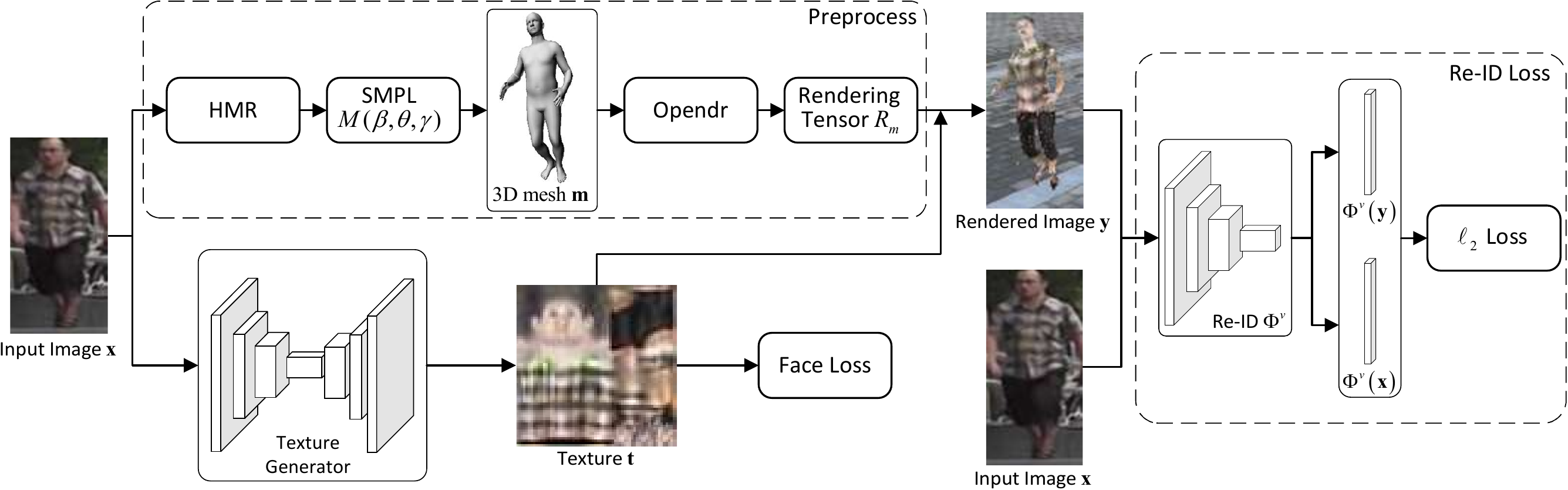}
	\caption{\textbf{Overview of the proposed framework.} Firstly, the image is sent to HMR and SMPL parameters are predicted. The 3D body mesh is calculated with HMR \cite{end2end} and SMPL \cite{smpl} and the rendering tensor is generated with Opendr. This step can be finished in the preprocessing procedure. Afterward, the texture is generated directly from the U-Net and the face loss of the generated texture is calculated. Finally, the rendered image $\mathbf{y}$ is generated as the product of the texture and rendering tensor. The background of $\mathbf{y}$ is randomly cropped from PRW dataset \cite{DBLP:conf/cvpr/ZhengZSCYT17} and CUHK-SYSU dataset \cite{xiao2016end}. The feature of the rendered image $\mathbf{y}$ and the input image $\mathbf{x}$ (denoted as $\Phi^v(\mathbf{x})$ and $\Phi^v(\mathbf{y})$ respectively) is extracted with a pretrained person re-identification model and the $\ell_2$ loss between $\Phi^v(\mathbf{x})$ and $\Phi^v(\mathbf{y})$ is calculated.}
	\label{method}
\end{figure*}

The key idea of our model is to maximize the perceptual similarity between the input images and the images rendered with generated textures. We suppose that the person rendered with better textures shares higher similarity with the person in input images. This is shown in Fig.~\ref{training} that similarity between the rendered person and input image is increasing along with the training process. 

The overall training procedure is depicted as follow. Firstly, we predict SMPL \cite{smpl} parameters with HMR \cite{end2end} and calculate the body mesh from predicted parameters (Sec.~\ref{mesh}). Subsequently, the texture generator, U-Net \cite{DBLP:conf/miccai/RonnebergerFB15}, is used to generate the texture from a single image. The differentiable renderer, Opendr \cite{DBLP:conf/eccv/LoperB14}, is further applied to render the human body image with the generated texture (Sec.~\ref{rendering}). After that, the input and rendered images are sent to a pretrained person re-identification network with part-based convolutional baseline \cite{pcb} and the distance between extracted features are minimized. In addition, to make the face of the generated texture more realistic, we also minimize the difference in the face parts between generated textures and the 3D scanned textures (Sec.~\ref{loss}). The overall architecture of our method, which is trained in an end-to-end way, is shown in Fig.~\ref{method}.

Recently, the generative adversarial network (GAN) \cite{goodfellow2014generative} has been widely used in image generation tasks \cite{pix2pix2017,DBLP:conf/nips/MaJSSTG17} as it can generate images that look superficially authentic to human observers. However, combining the loss with a GAN-style discriminator will not work in our method. As there is an obvious style gap between the rendered images and real ones, the discriminator in GAN can always distinguish them easily, which causes the gradient of generator diminishes. 

\subsection{Body Mesh Reconstruction}\label{mesh}
In our method, we render our textures on the SMPL body model due to its outstanding realism and high computational efficiency. SMPL parameterizes human body mesh with shape parameters $\beta \in \mathbb{R}^{10}$, pose parameters $\theta \in \mathbb{R}^{72}$ and translation parameters $\gamma \in \mathbb{R}^3$. The shape parameters control how individuals vary in height, weight and body proportions, while the pose parameters model the 3D rotation of both the human body and the $K=23$ joints in axis-angle representation. The translation parameters are optional as it controls the position of the human body mesh in the orthogonal coordinate system. SMPL uses a differentiable function $M(\theta,\beta, \gamma) \in \mathbb{R}^{3\times N}$ to give the triangulated body mesh with $N=6890$ vertices.

Although the re-identification network can reduce the influence caused by variations in body pose and translation, these variations, especially the position and orientation of the human body, can still interfere with the training process, which is shown in Sec.~\ref{ablation}. Thus, we still need to align the rendered person with the input image by estimating body shape, pose, and translation of the input image. To tackle this issue, we adopt HMR, the state-of-the-art method for the 3D body pose and shape estimation. HMR produces the shape, pose and translation parameters for SMPL with an iterative 3D regression module. Thus, the estimated 3D mesh $\mathbf{m}$ from the image $\mathbf{x}$ can be expressed as follows: $\mathbf{m} = M(\beta,\theta,\gamma)=M(hmr(\mathbf{x}))$.

\subsection{Texture Rendering}\label{rendering}

In this step, we generate the texture with U-Net and apply Opendr \cite{DBLP:conf/eccv/LoperB14}, a differentiable renderer, to map the generated texture to the 3D mesh. With the UV correspondence provided by SMPL, the rendering function $R(\mathbf{m}, \mathbf{t})$ of Opendr directly assigns pixels to surface on the 3D mesh polygon and fills in the gaps with linear interpolation. 

With the fixed 3D human mesh $\mathbf{m}$, the rendering function $R(\mathbf{m}, \mathbf{t})$ can be viewed as a linear transformation that maps from the space of texture $\mathbf{t}$ to the space of rendered image $\mathbf{y}$:
\begin{equation}
R_{\mathbf{m}} = \mathbf{R}_{h_{\mathbf{t}} \times w_{\mathbf{t}}\times c\times h_{\mathbf{y}}\times w_{\mathbf{y}}\times c}
\end{equation}
where $h_{\mathbf{t}}$ and $w_{\mathbf{t}}$ stand for the height and width of texture image, $h_{\mathbf{y}}$ and $w_{\mathbf{y}}$ stand for the height and width of rendered image and $c$ stands for the size of image channels.

The rendering process can be simplified as the multiplication of tensors:
\begin{equation}
\mathbf{y} = R_{\mathbf{m}}(\mathbf{t}) = \mathbf{t} \otimes R_{\mathbf{m}}
\end{equation}

The rendering tensor $R_{\mathbf{m}}$ will not change as long as the human 3D mesh is fixed. This provides a trick for accelerating the training procedure: we can predict all 3D meshes and calculate all rendering matrices $R_{\mathbf{m}}$ of training data in the preprocess step. In this way, we can avoid the time-consuming process of calculating rendering tensor $R_{\mathbf{m}}$ in the training phase.

\subsection{Loss Functions}\label{loss}

\paragraph{Re-identification loss.}\label{reid-loss}
The re-identification loss is the layer-wise feature distance between rendered image and input image. Given a pair of input and rendered image $\{\mathbf{x},\mathbf{y}\}$, we use the pretrained re-identification network as a feature extractor for both $\mathbf{x},\mathbf{y}$. We penalize the $\ell_2$ distance of the respective intermediate feature activations $\Phi^v$ at $n = 4$ different network layers ($v=1,...,n$) after the Resnet block.

\begin{equation}
\mathcal{L}_{reid}(\mathbf{x},\mathbf{y})=\sum_{v=1}^{n}\Vert \Phi^v(\mathbf{x}) - \Phi^v(\mathbf{y}) \Vert_2
\end{equation}

This loss penalizes differences in low- mid- and high-level feature statistics, captured by respective network filters. 


The setting of re-identification loss is similar to the perceptual loss which is widely used for image generation while the perceptual loss uses a network pretrained on ImageNet. However, our method outperforms the model trained on perceptual loss, which will be shown in Sec.~\ref{ablation}. This is because the re-identification network has been explicitly trained to minimize the distance of the images of the same person and maximize that of the different persons. As the person identity is mostly characterized by the body texture, the re-identification network performs better for guiding the texture generation process.

In our proposed approach, we use the person re-identification model with PCB \cite{pcb} because of its simplicity and efficiency in extracting features from different body parts. Other re-identification models can reach similar results while they perform badly when generating the details of the human body, which will be shown in Sec.~\ref{ablation}.

\paragraph{Face loss.}

In order to improve the realism of generated texture, we design the face loss as the $\ell_1$ loss of face and hand parts between the generated texture $\mathbf{t}$ and 3D scanned texture $\mathbf{t_s}$ from SURREAL \cite{synthetichumans}. Given the mask $\mathcal{M}$ of head and hand parts, the face loss is defined in the following way:

\begin{equation}
\mathcal{L}_{face}(\mathbf{t},\mathbf{t_s})=\Vert \mathcal{M} \odot (\mathbf{t} - \mathbf{t_s}) \Vert_1
\end{equation}

\begin{figure}[h]
	\centering
	\begin{subfigure}[c]{0.22\columnwidth}
		\includegraphics[width=\textwidth]{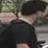}
		\caption*{Input}
	\end{subfigure}
	\hspace{0.05em}
	\begin{subfigure}[c]{0.22\columnwidth}
		\includegraphics[width=\textwidth]{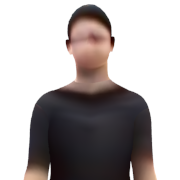}
		\caption*{No $\mathcal{L}_{face}$}
	\end{subfigure}
	\hspace{0.05em}
	\begin{subfigure}[c]{0.22\columnwidth}
		\includegraphics[width=\textwidth]{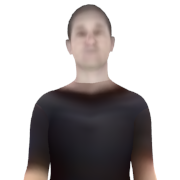}
		\caption*{Pasted face}
	\end{subfigure}
	\hspace{0.05em}
	\begin{subfigure}[c]{0.22\columnwidth}
		\includegraphics[width=\textwidth]{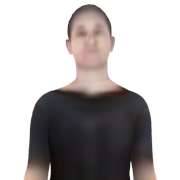}
		\caption*{Full method}
	\end{subfigure}
	\caption{\textbf{Results with and without face loss.} For ``Pasted face'', we copy the face part of textures from SURREAL and paste it to the texture of ``No $\mathcal{L}_{face}$''.}
	\label{face_loss}
\end{figure}

The face loss makes the face part of the generated texture similar to the corresponding part in the scanned texture. The reason why we use the face loss in the training procedure rather than simply covering the generated face parts with scanned ones is that the former approach can eliminate the color contrast between head and torso. From Fig.~\ref{face_loss} we can see the model trained without face loss produces results of low quality. If we substitute the face part with textures in SURREAL, there will be an obvious color contrast on the neck.

In all, our overall loss function is:
\begin{equation}
\mathcal{L} = \lambda_{reid}\mathcal{L}_{reid} + \lambda_{face}\mathcal{L}_{face}
\end{equation}
where the $\lambda_{reid}$ and $\lambda_{face}$ are the weight of re-identification loss and face loss respectively.

\section{Experiments}
\subsection{Datasets and Metrics}\label{datasets_and_metrics}
\paragraph{Datasets.}
We perform our experiments on commonly-used re-identification dataset Market-1501 \cite{DBLP:conf/iccv/ZhengSTWWT15}, containing 32,668 images of 1,501 identities captured from six disjoint surveillance cameras. All
images are resized to $64 \times 128$ pixels. In our experiment, we select 100 person identities for testing and the remaining 1401 for training. We also removed all the images with unknown human labels. This results in 30,470 training and 1,747 testing images.


\paragraph{Metrics.}
Evaluating the quality of image generation method is a tricky task. In our experiments, we adopt a redundancy of metrics and a user study to evaluate the quality of generated textures. Following \cite{DBLP:conf/nips/MaJSSTG17}, we use the Structural Similarity (SSIM) \cite{DBLP:journals/tip/WangBSS04}, Inception Score (IS) \cite{DBLP:conf/nips/SalimansGZCRCC16} and the masked version of them: mask-SSIM and mask-IS \cite{DBLP:conf/nips/MaJSSTG17}.

The mask-SSIM is incorporated in order to reduce the influence of background in our evaluation. A pose mask is added to both the generated and the target image before computing SSIM. In this way, we only focus on measuring the generation quality of a person's appearance.

Though the SSIM performs well in estimating similarity both in body structure and textures, the Inception Score is not useful in our problem. This is because it only rewards the inter-class divergence and penalizes the inner-class divergence, which means that it is not relevant to the with-in class object generation \cite{neverova2018dense}. We also have empirically observed its instability with respect to the perceived quality and structural similarity. Thus, we do not expect to draw strong conclusions from it.

\subsection{Implementation Details}
We train texture generator with the Adam optimizer (learning rate: $2 \times 10^{-4}$, $\beta_1 = 0.9$, $\beta_2 = 0.999$, weight decay: $1 \times 10^{-5}$). In
all experiments, the batch size is set to 16 and the training proceeds for 120 epochs, totally 64k iterations. Every batch contains four groups while each group constitutes of four images from the same person identity.
The balancing weights $\lambda$ between different losses (described
in Sec.~\ref{loss}) are set empirically to $\lambda_{reid}=5\times10^3$, $\lambda_{face}=1.0$.

\subsection{Comparison with Available Methods}\label{comparison_with_available}
\begin{figure}[h]
	\centering
	\includegraphics[width=0.97\linewidth]{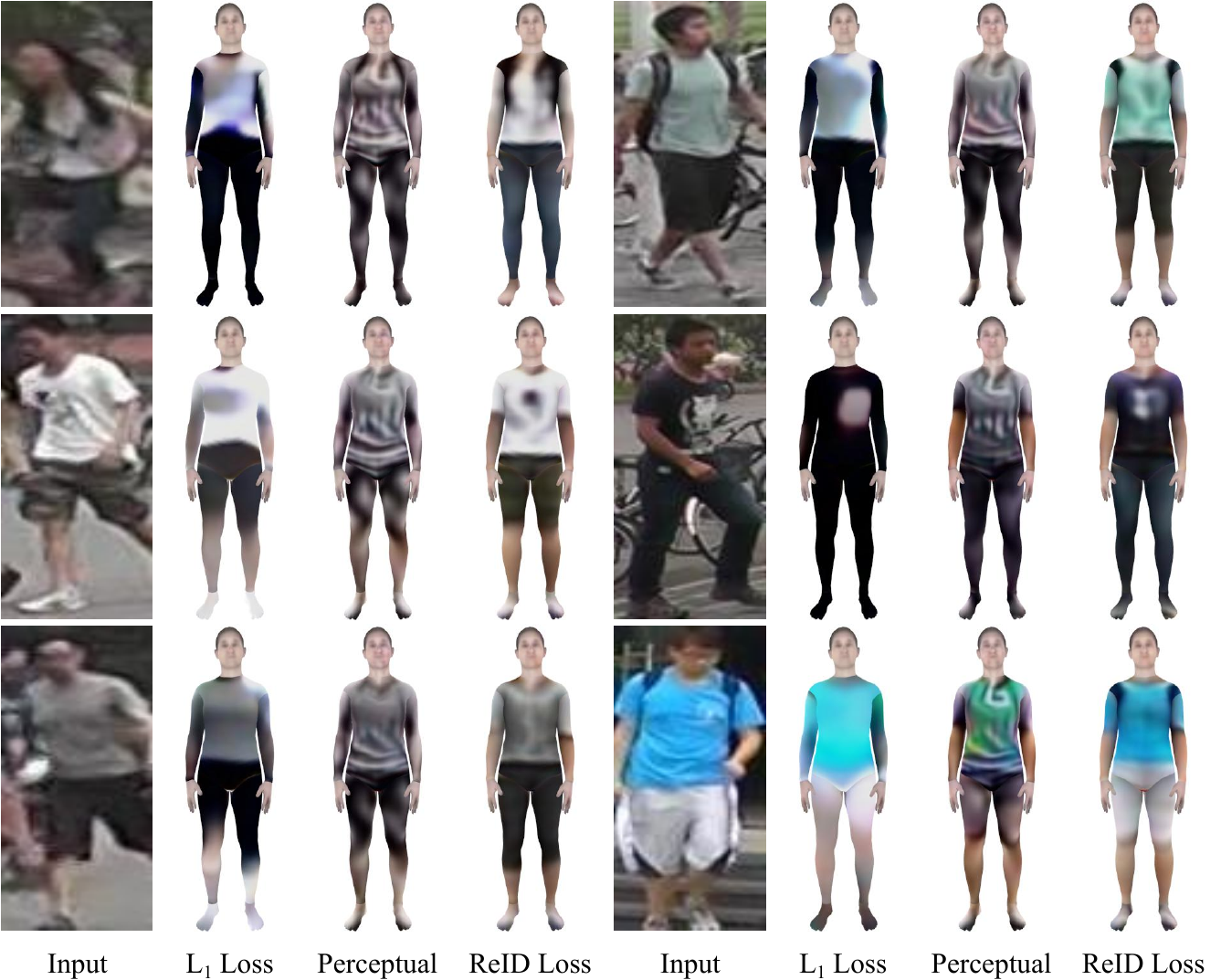}
	\caption{\textbf{Qualitative results.} The result of different loss functions is shown above. Each column shows (in order from left): input images (from test set), pixel-wise $\ell_1$ loss, perceptual loss, the re-identification loss}
	\label{result_img}
\end{figure}

In this section, we demonstrate that the re-identification plays an indispensable role in our method. We compare the re-identification loss in our approach with two commonly-used loss functions for the texture generation task: the pixel-wise $\ell_1$ loss \cite{neverova2018dense} and perceptual loss \cite{DBLP:conf/eccv/KanazawaTEM18}. We show our qualitative results in Fig.~\ref{result_img} and quantitative results in Table.~\ref{result_table}. From the qualitative result, we can conclude that the model trained with re-identification loss performs better than models with other loss functions. The reason why the pixel-wise $\ell_1$ and perceptual loss performs bad is analyzed in the following paragraphs.

The pixel-wise $\ell_1$ loss is defined as the $\ell_1$ loss between the rendered and input image. The model trained with pixel-wise $\ell_1$ reconstruction loss performs bad especially in generating details, such as hands and shins, of the image. This is caused by the inaccurate estimation of human body poses and shapes provided by HMR \cite{end2end} module. 

\begin{table}[h]
	\begin{center}
		\begin{tabular}{l c c c}
			\hline
			Model & pixel-wise $\ell_1$& Perceptual&ReID Loss\\
			\hline
			SSIM & 0.162 & 0.149 &\textbf{0.164}\\
			mask-SSIM & \textbf{0.374} & 0.356& 0.372\\
			\hline
		\end{tabular}
	\end{center}
	\caption{\textbf{Quantitative results.} The SSIM of our method is higher than others while our mask-SSIM score is only 0.02 less than results of pixel-wise $\ell_1$ loss.}
	\label{result_table}
	
\end{table}

As described in \cite{DBLP:conf/iccv/ChenK17}, the perceptual loss is defined as the $\ell_2$ distance between features of two images extracted from 5 intermediate layers of a VGG19 network which is pretrained on ImageNet. The model trained with perceptual loss ended up with an even worse result. The perceptual quality of the torso part is poor. This is because the network trained on Imagenet tends to extract general features of objects rather than concentrating on body textures. 

The SSIM and mask-SSIM score of $\ell_1$ reconstruction loss are among the highest scores in all of the experiment results, this is because the $\ell_1$ loss optimize the generated texture in a pixel-to-pixel way, which is equivalent to directly optimizing the SSIM score in the training process. While the re-ID loss does not directly optimize the SSIM score, the SSIM score is still high using re-ID loss. This verifies that re-ID loss is indeed effective.

The IS scores of the three models are 3.96, 4.04 and 3.96 respectively while the mask-IS scores are 2.90, 2.59 and 2.52. We do not show these results in Table~\ref{result_table} for the same reason in Sec.~\ref{datasets_and_metrics}.

\subsection{Ablation Study}\label{ablation}

\begin{figure}[h]
	\centering
	\includegraphics[width=0.97\linewidth]{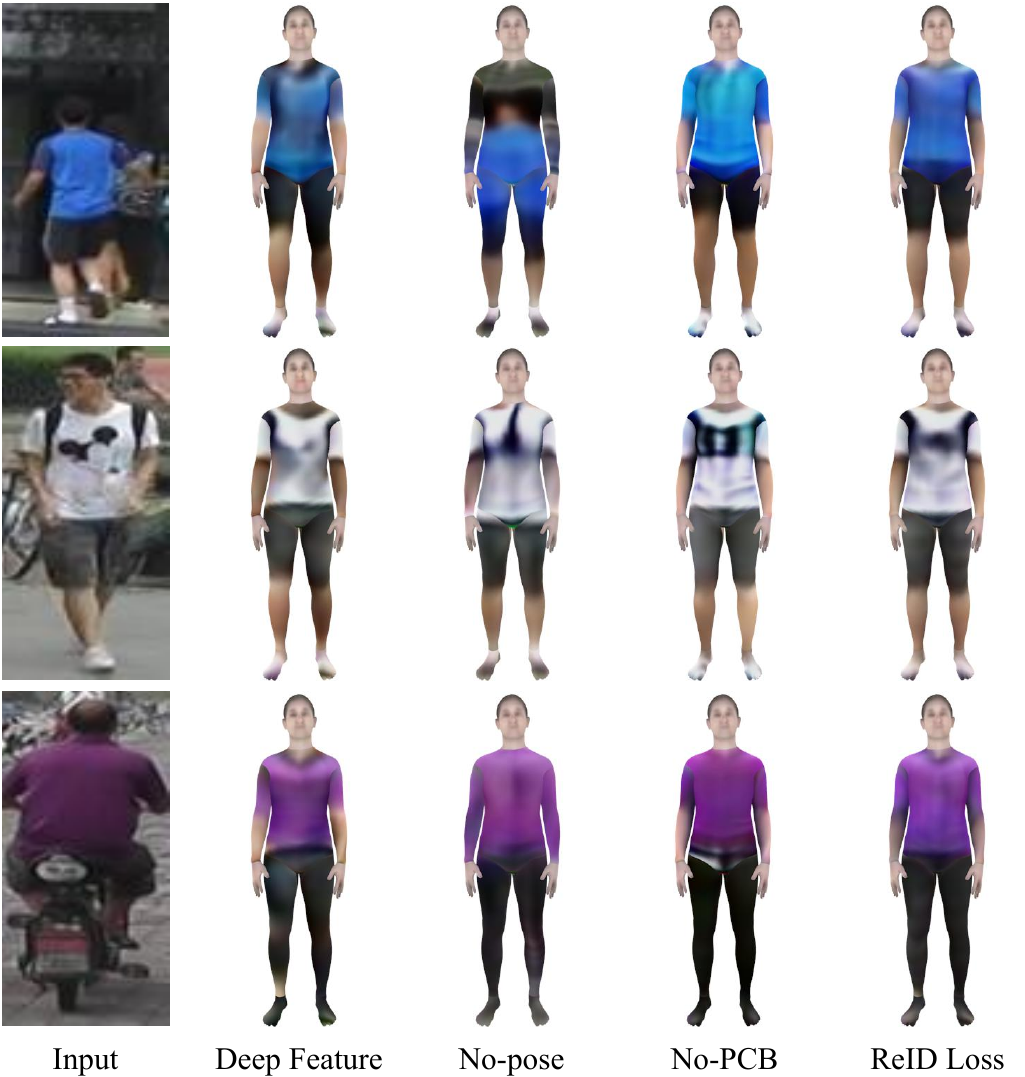}
	\caption{\textbf{Qualitative results of ablation study.} The result of different model settings is shown above. Each column shows (in order from left): input images (from test set), the $\ell_1$ loss on deep features, model without the pose alignment, re-identification loss without body part features, the re-identification loss proposed in Sec.~\ref{reid-loss}}
	\label{ablation_img}
\end{figure}

In this section, we carry out the following experiments to explore the influence of different model settings. The qualitative result is shown in Fig.~\ref{ablation_img} while the quantitative result is shown in Table.~\ref{ablation_table}.

\paragraph{$\ell_1$ loss of deep features.}
In the re-identification network with PCB, the image is passed into a resnet-50 network and a pooling layer, producing the feature $g$ of $6 \times 256$ dimensions. The deep feature loss is defined as the $\ell_1$ distance between deep features $g$ of re-identification network. 
%

The result of deep feature loss is shown in the column labeled ``Deep Feature'' of Fig.~\ref{ablation_img}. Compared with the proposed re-identification loss, this result is good while ignoring some details, e.g. the patterns on clothes. This is because the deep features can hardly represent the details of human texture as the deep features $g$ can hold less information than features from different layers. The qualitative result is also confirmed by the SSIM and mask-SSIM scores.

\paragraph{With vs. without body pose alignment.}

In our method, we estimate the body pose and shape parameters of the images with HMR and render the SMPL body model with these parameters. This can be viewed as the alignment of pose and position of human body. Even though the re-identification is believed to own the ability of filtering out the influence of body posture and position, we still suppose that the body pose alignment contributes to the texture generation process because the influence of body pose and position is inevitable \cite{alignedreid}.

In this experiment, we substitute the SMPL parameters with the randomly chosen parameter from the walking sequences of the CMU MoCap database \cite{cmumocap}. The result is shown in the column labeled ``No-pose'' of Fig.~\ref{ablation_img}. The experiment shows that the model without pose alignment can generate textures of acceptable quality. However, the defects in arm parts indicates the significance of body alignment. Moreover, in the first example of Fig.~\ref{ablation_img} where human only occupies half part of the image, the model without pose alignment cannot recovery the human body size and location automatically and it regards the background as a part of the body.

The SSIM and mask-SSIM scores of the model without pose alignment are 0.158 and 0.365 respectively, which are lower than our method. This result indicates that the feature extracted by the re-identification model more or less influenced by the pose and position of the human body, which is consistent with the conclusion in \cite{alignedreid}.

\paragraph{With vs. without body part features.}

In our method, we employ PCB model \cite{pcb} as the feature extraction module of re-identification loss because it can extract features from every part of the human body, which is supposed to be beneficial for reconstructing details of human body. To demonstrate this, we compared the performance of our method with the model without body part features. In order to exclude the influence caused by the different performances of re-identification network, the top-1 accuracy of the two re-id networks are both around 92\% on the Market-1501 dataset.

The result is shown in the column labeled ``No-PCB'' of Fig.~\ref{ablation_img}. The result with and without body part features is mostly similar while they only differs on the arms or clothing details. This aligns well with the SSIM and mask-SSIM results, where the scores of No-PCB is slightly lower than proposed method. The high quality textures generated by the model without PCB can be ascribed to the high accuracy of re-identification model which extracts precise features.

\begin{table}[h]
	\begin{center}
		\setlength{\tabcolsep}{0.8mm}{
		\begin{tabular}{l c c c c}
			\hline
			Model & Deep Feature&No-Pose&No-PCB&ReID Loss\\
			\hline
			SSIM& 0.155& 0.158& 0.159& \textbf{0.164}\\
			mask-SSIM & 0.354& 0.365& 0.369& \textbf{0.372}\\
			\hline
		\end{tabular}}
	\end{center}
	\caption{\textbf{Quantitative results of ablation study.} The SSIM and mask-SSIM score of ReID loss is the higher than other loss functions.}
	\label{ablation_table}
	
\end{table}

The IS scores of the four methods are 3.77, 4.07, 3.75 and 3.96 respectively while the mask-IS scores are 2.37, 2.63, 2.77 and 2.52. We do not show these results in Table~\ref{ablation_table} for the reason in Sec.~\ref{datasets_and_metrics}.

\subsection{User Study}

A commonly used way to further assess the reality of generated texture is the user study, as human judgment is the ultimate criterion in the generative model. However, unlike previous works \cite{DBLP:conf/cvpr/IsolaZZE17}, our network generates the textures instead of images of human, which makes it impossible for an unprofessional user to tell which texture is better. Moreover, the available rendering software cannot bridge the style gap between rendered and real images, which makes a direct comparison between them impossible. To tackle this issue, we designed the user study aiming at comparing the generated textures with 3D-scanned textures which can be considered as the ``real image'' in the domain of texture. We generated 55 image pairs and each pair contains one image rendered with the generated texture and another one rendered with the real texture from SURREAL~\cite{synthetichumans}. 30 users have to choose one image with higher quality among two images in 2 seconds. The first 5 image pairs are used for practice thus are ignored when computing scores. 

From the result of our user study, users consider the quality of generated textures is higher than scanned textures in 32\% image pairs. This shows the relatively high quality of textures generated by our method. By reviewing our user study, we find that the generated textures suffer from blurring while the 3D scanning tends to preserve the details. There are two reasons behind this. Firstly, our textures are generated from blurred images in Market-1501 dataset. Secondly, the differential render, Opendr, only performs well on textures of small size, which limits the resolution of our generated textures.

\subsection{Bird Texture Generation}\label{bird_sec}
Apart from generating textures of the human body, our framework can also be applied to other object categories. \cite{DBLP:conf/eccv/KanazawaTEM18} presents a learning framework called CMR for recovering the 3D shape, camera, and texture of an object from a single image. CMR projects the 3D mesh to 2D images and uses the perceptual loss~\cite{zhang2018perceptual} between the rendered image and input image as the loss function. We believe our re-identification supervised method can outperform CMR as our loss function performs better than the perceptual loss, which is demonstrated in Sec.~\ref{comparison_with_available}.

\begin{figure}[h]
	\centering
	\begin{minipage}[b]{0.9\columnwidth}
		\begin{subfigure}[c]{0.32\columnwidth}
			\includegraphics[width=\textwidth]{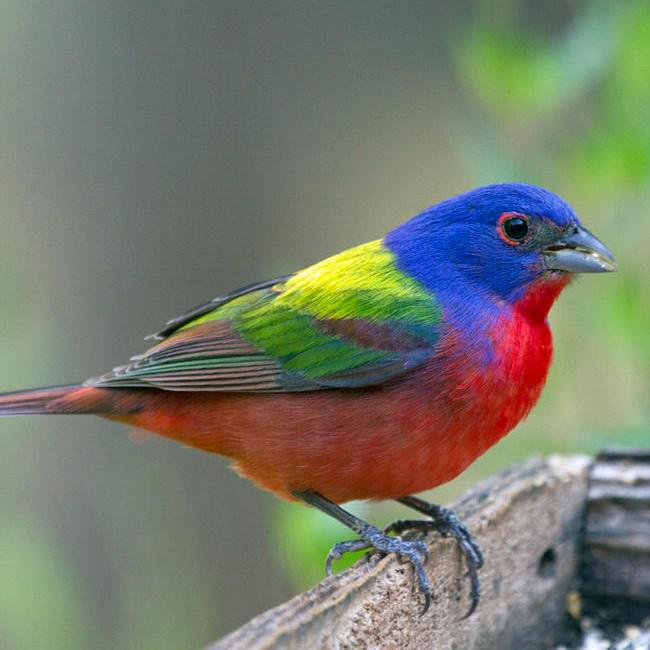}
			\label{fig:1}
		\end{subfigure}
		\hfill 
		\begin{subfigure}[c]{0.32\columnwidth}
			\includegraphics[width=\textwidth]{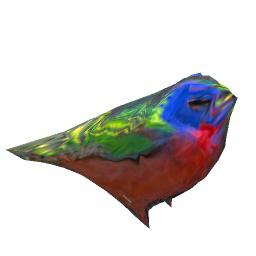}
			\label{fig:2}
		\end{subfigure}
		\hfill 
		\begin{subfigure}[c]{0.32\columnwidth}
			\includegraphics[width=\textwidth]{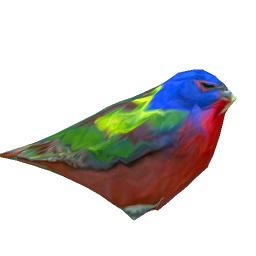}
		\end{subfigure}
		\vspace{-1em}
	\end{minipage}
	\begin{minipage}[b]{0.9\columnwidth}
		\begin{subfigure}[c]{0.32\columnwidth}
			\includegraphics[width=\textwidth]{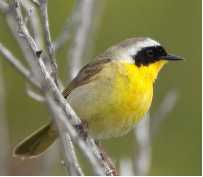}
			\caption*{Input}
			\label{fig:1}
		\end{subfigure}
		\hfill
		\begin{subfigure}[c]{0.32\columnwidth}
			\includegraphics[width=\textwidth]{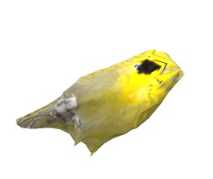}
			\caption*{CMR~\cite{DBLP:conf/eccv/KanazawaTEM18}}
			\label{fig:2}
		\end{subfigure}
		\hfill 
		\begin{subfigure}[c]{0.32\columnwidth}
			\includegraphics[width=\textwidth]{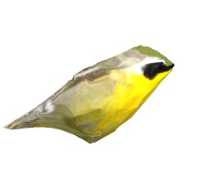}
			\caption*{Ours}
			\label{fig:2}
		\end{subfigure}
	\end{minipage}
	\caption{\textbf{Sample results.} The textures generated by our method is of higher quality than textures of CMR. This is clearly shown on the heads of birds.}
	\label{birds}
\end{figure}

To compare with them, we trained the re-identification network on CUB-200-2011 dataset \cite{WahCUB_200_2011} and use it as the perceptual metrics in the texture generation module. The CUB-200-2011 dataset has 6,000 training and 6,000 test images from 200 birds species. Following \cite{DBLP:conf/eccv/KanazawaTEM18}, we filter out nearly 300 images where the visible number of keypoints are less than or equal to 6. Our experiment result is shown in Fig.~\ref{birds}.

%

From the experiment results we can conclude that the textures generated with re-identification loss are more accurate than those generated with the perceptual loss. Particularly, our method succeeds in reconstructing details of the input image. For example, the bird heads of our generated textures look more realistic than CMR \cite{DBLP:conf/eccv/KanazawaTEM18}. This experiment shows that our method holds the potential to be applied in the texture generation task for general categories.

\section{Action Recognition}
SURREAL \cite{synthetichumans} provides an effective approach to generate the synthetic dataset and proves that pretraining on synthetic dataset can enhance the performance of human parsing and depth estimation models. However, due to the shortage of available human body textures, the generated dataset suffers from a lack of texture diversity. This limits the generalize ability of models pretrained on the synthetic datasets.

Our method provides an efficient way to generate a large number of textures, which can be used to synthesize datasets with higher diversity and tackle the aforementioned issue. In order to prove this, we carry out the experiment to compare the networks pretrained on action recognition datasets synthesized with different textures. One dataset is the SURREAL dataset generated with 772 scanned textures while another dataset called SURREAL++ is generated with 1.5k textures extracted from Market-1501. We generate SURREAL++ with the method proposed in SURREAL using sequences of 2607 action categories from CMU MoCap dataset \cite{cmumocap}. This makes the SURREAL++ embody 67,582 continuous image sequences containing 6.5 million frames, which is of the same size as the SURREAL dataset.

To evaluate the generated dataset, we implement the non-local neural networks \cite{NonLocal2018} which is commonly used for action recognition task and pretrain it on both the SURREAL and SURREAL++ dataset. Then we fine-tune the networks and test them on UCF101 dataset \cite{soomro2012ucf101} to estimate the networks' performance. The UCF101 dataset contains 13320 videos from 101 action categories. It uses three train/test splits and each split contains around 9.5k training videos and 3.7k test videos.  We report our method by the average of 3-fold cross-validation. In addition, we use the non-local network trained on UCF101 dataset as our baseline model.

\begin{table}[h]
	\begin{center}
		\setlength{\tabcolsep}{2mm}{
		\begin{tabular}{l c c}
			\hline
			Training Data  & Top-1 Acc. (\%) & Top-5 Acc. (\%)\\
			\hline
			UCF101 (baseline)&82.03&94.43\\
			SURREAL&85.83&96.98\\
			SURREAL++&\textbf{86.89}&\textbf{97.04}\\
			\hline
		\end{tabular}
	}
	\end{center}
	\caption{\textbf{Experiment results.} We show the top-1 and top-5 accuracy of models trained on different datasets.}
	\label{action}
	
\end{table}
Table.~\ref{action} summarizes test results on UCF101. The model pretrained on the SURREAL dataset and fine-tuned on UCF101 is 3.80\% higher in top-1 accuracy than the baseline model, while the model pretrained on SURREAL++ dataset is 4.86\% higher in top-1 accuracy. This result shows that the dataset with richer texture diversity can elevate the generalize ability of networks and such kind of diversity can be obtained with our method.

\section{Conclusion and Future Work}
In this paper, we present an end-to-end framework for generating the texture from a single RGB image. This is achieved by incorporating the pretrained re-identification network as the supervision for texture generation. We have shown that re-identification network can work as a good supervisor in the texture generation task due to its ability to extract body features while reducing the influence in pose variations. We have also proved the extensive application potential of re-identification network in the 3D reconstruction of general categories. To provide the possible usage of our generated body textures, we have demonstrated the diversity in our textures can provide the pretrained model with higher performance.

As the quality of the generated human body texture is restricted by low-quality differentiable render, we suppose that a high-quality renderer will enhance the performance of our method dramatically. We also note that as our framework renders a synthetic image in a similar pose as that of the input image, the quality of texture in occluded parts is not guaranteed. However, from the training images, we can find another image $\mathbf{x'}$ with the same identity as the input image$\mathbf{x}$, but in a different pose. Then, we can align the pose of rendered image $\mathbf{y}$ with $\mathbf{x'}$ and therefore supervise the texture generation process under another viewpoint. This extension will be explored in the future.

\textbf{Acknowledgement.} This research was supported by National Key R\&D Program of China (No. 2017YFA0700800).

{\small
\bibliographystyle{ieee}
\bibliography{references}
}

\end{document}